\documentclass{article}
\usepackage{spconf,amsmath,graphicx,bm}

\title{Deep generative LDA}
\name{Yunqi Cai, Dong Wang*}
\address{Department of Computer Science and Technology,\\
         Center for Speech and Language Technologies,\\
         Tsinghua University, Beijing, China\\
        }

\begin{document}
%
\maketitle
\begin{abstract}

Linear discriminant analysis (LDA) is a popular tool for classification and dimension reduction. Limited by its linear form and the underlying Gaussian assumption, however, 
LDA is not applicable in situations where the data distribution is complex. Recently, we proposed a discriminative normalization flow (DNF) model.
In this study, we reinterpret DNF as a deep generative LDA model, and study its properties in representing complex data. 
We conducted a simulation experiment and a speaker recognition experiment. The results show that DNF and its subspace version are
much more powerful than the conventional LDA in modeling complex data and retrieving low-dimensional representations.

\end{abstract}

\section{Introduction}
\begin{figure}[htb]
    \centering
    \includegraphics[width=0.85\linewidth]{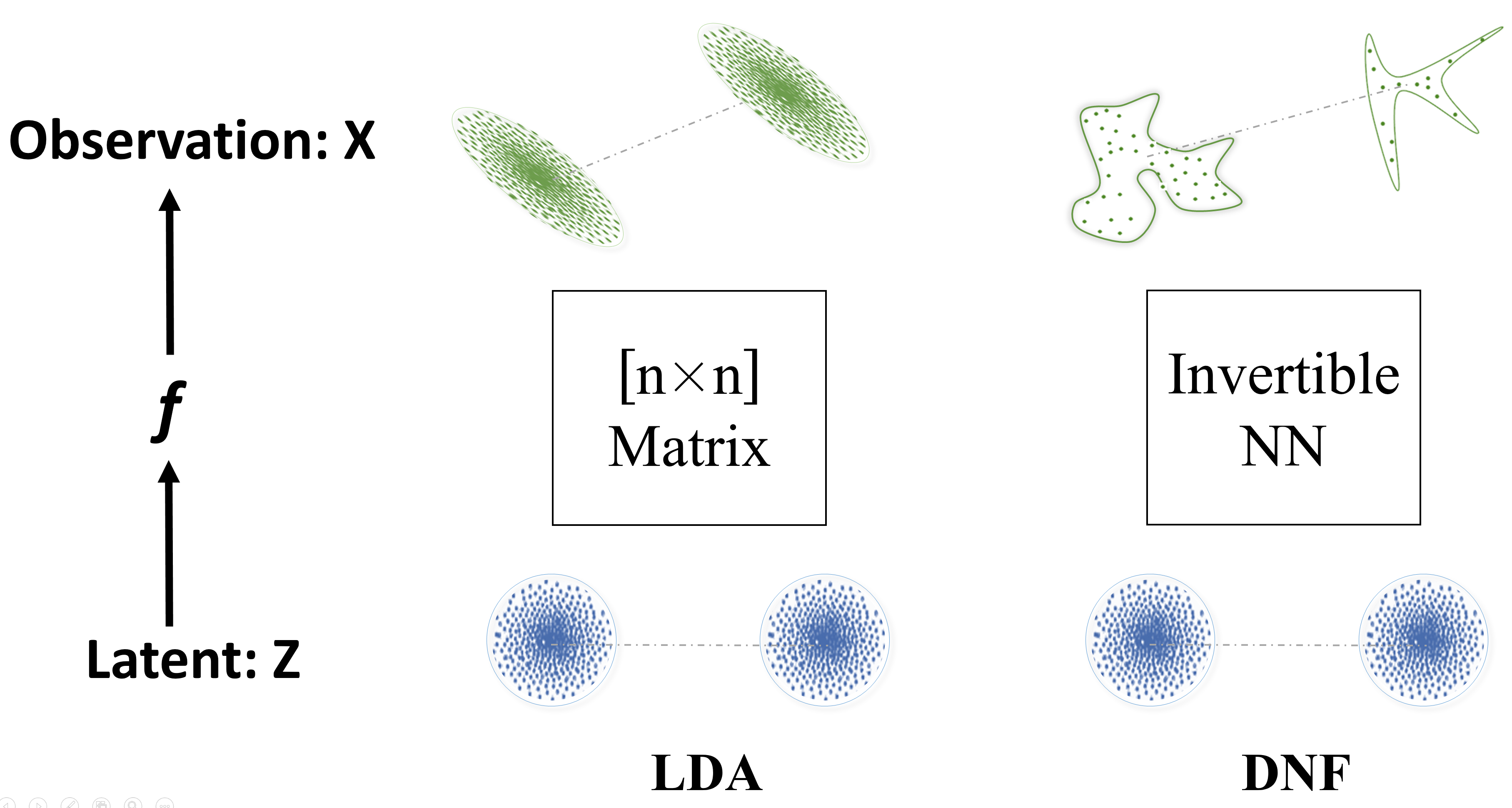}
    \vspace{-4mm}
    \caption{(Left) LDA maps standard Gaussians to homogeneous Gaussians by a linear transformation; (Right) DNF maps 
    standard Gaussians to heterogeneous distributions in complex forms.}
    \label{fig:dglda}
\end{figure}

\vspace{-3mm}
\label{sec:intro}
{\setlength{\parskip}{1.2em}

Linear discriminant analysis (LDA), sometimes called Fisher discriminant analysis (FDA), was first proposed by Fisher in 1936~\cite{fisher1936use} 
and is a classic tool for dimension reduction. 
Although the original form is purely discriminative,
Campbell et al.~\cite{campbell1984canonical} showed that LDA corresponds to a linear Gaussian model, 
where each class is represented by a Gaussian distribution, and different classes share the same covariance matrix. 
This model is purely generative and is trained by maximum likelihood (ML). By this generative model view, 
LDA is regarded as a general model for discrimination tasks, and dimension reduction is just a by-product 
and can be derived by constraining the class means in a subspace~\cite{campbell1984canonical}.

In spite of the wide application, blindly employing LDA may suffer from serious problems if the data is not distributed as 
homogeneous (covariance-shared) Gaussians,
e.g., the small sample size (SSS) problem~\cite{sharma2015linear,belhumeur1997eigenfaces}, 
non-linearly separate problem~\cite{sharma2015linear, scholkopft1999fisher}, and heteroscedasticity and non-Gaussianity 
problem~\cite{campbell1984canonical, engle1995arch,kumar1998heteroscedastic}. In order to handle complex data, the homogeneous 
Gaussian assumption has to be relaxed. To mention several, the quadratic discriminant analysis (QDA)~\cite{tharwat2016linear}
and the heteroscedastic LDA (HLDA)~\cite{kumar1998heteroscedastic} relax the shared-covariance assumption; 
the mixture discriminant analysis (MDA)~\cite{hastie1996discriminant} relaxes the Gaussian assumption by 
allowing data to distribute as a Gaussian mixture;
the kernel discriminant analysis~\cite{mika1999fisher} relaxes the Gaussian assumption by introducing a non-linear transformation. 
The above extensions (and other alternatives) can deal with non-Gaussian and non-homogeneous data, but 
their learning capacity is often constrained by their model forms and so cannot represent \emph{general} data. 
Moreover, some of the extensions are only applicable for discrimination tasks, but not for dimension reduction, e.g., 
QDA~\cite{tharwat2016linear}. 

Deep neural networks (DNN) have shown remarkable power in learning abstract features from complex data~\cite{goodfellow2016deep}.
From the perspective of dimension reduction, DNN can perform this task extremely well by compressing class information 
into a short feature vector, by performing discriminative training with the cross-entropy objective, or an objective that is 
exactly the same as LDA, the Fisher criterion~\cite{dorfer2015deep}. However, the features derived by the
discriminative training will not gain any simple form of distribution, and generally lose their capacity of explaining the data.

Recently, we proposed a discriminative normalization flow (DNF) model and used it to normalize class distributions~\cite{cai2020deep}. 
In this paper, we reinterpret DNF as a nonlinear extension of LDA. Compared to the early work such as QDA and
MDA, the new model is based on deep neural networks and so inherits their brilliant power in learning from data, which grants DNF the capacity of modeling 
any kind of data in theory. Compared to other `deep LDA' models that are based on DNNs, our model is generative and can explain the data by generation. 
Most importantly, our model holds the homogeneous Gaussian assumption in the latent space, and can be trained by 
maximum likelihood. All of these features make it highly resemble the conventional LDA, except that the linear transformation is
replaced by a neural networks. When the DNN is implemented as a linear transformation, the model falls back to the standard LDA. Due to 
such resemblance, DNF can be interpreted as \textbf{deep generative LDA}, to emphasize its deep nature and discriminate it 
from other `deep LDAs' based on discriminative training~\cite{dorfer2015deep}. Fig.~\ref{fig:dglda} compares the LDA
and DNF.
}

\vspace{-5mm}

\section{Deep LDA by Discriminative Normalization Flow}

\subsection{Probabilistic view of LDA}


LDA can be cast to a linear Gaussian model~\cite{campbell1984canonical}, where each class is a Gaussian and all classes share the same covariance, as formulated below:

\begin{equation}
\label{eq:LDA-prior-x}
p(\mathbf{x}; y(\mathbf{x})) =  N(\mathbf{x}; \pmb{\nu}_{y(\mathbf{x})}, \pmb{\Sigma}).
\end{equation}

\noindent Define an invertible linear transformation $\mathbf{z}=\mathbf{M}\mathbf{x}$ that projects the observation $\mathbf{x}$ to a latent 
code $\mathbf{z}$, for which each class is a standard Gaussian. 

\begin{equation}
\label{eq:LDA-prior-z}
p(\mathbf{z};y(\mathbf{x})) =  N(\mathbf{z}; \pmb{\mu}_{y(\mathbf{x})}, \pmb{I}),
\end{equation}

\noindent The probability $p(\mathbf{x})$ and the probability $p(\mathbf{z})$ has the following relation: 

\begin{equation}
\label{eq:ML-LDA}
 p(\mathbf{x};y(\mathbf{x})) = p (\mathbf{z};y(\mathbf{x})) |\det \mathbf{M}|.
\end{equation}

\noindent \noindent where $\det \mathbf{M}$ represents determinant of $\mathbf{M}$. Given a set of training data $\{\mathbf{x}_i\}$, 
the likelihood of the data can be computed as follows:

\begin{eqnarray}
L(\mathbf{M}) &=& \sum_i \log (p(\mathbf{x}_i; y(\mathbf{x}_i))) \\
&=&  \sum_i \big\{ \log (p(\mathbf{z}_i; y(\mathbf{x}_i))) + \log |\det\mathbf{M}| \big\}
\end{eqnarray}

\noindent Maximizing $L(\mathbf{M})$ with respect to $\mathbf{M}$ leads to the full-dimension LDA.

As shown by Kumar et al.~\cite{kumar1998heteroscedastic}, LDA-based dimension reduction can be derived by constraining the class means 
in a low-dimensional subspace. For that purpose, one can assume that only part of the dimensions of the class means in the latent space 
are class-dependent, as shown below:
 
\begin{equation}
\label{eq:split}
\pmb{\mu}_y = \begin{bmatrix}
\pmb{\mu}_y^c\\
\pmb{\mu}^0
\end{bmatrix},
\end{equation}

\noindent where $\pmb{\mu}_y^c$ is class-dependent and $\pmb{\mu}^0_y$ is shared by all classes. Under this assumption, optimizing the transformation $\mathbf{M}$ via maximum 
likelihood leads to the dimension-reduction LDA, and the submatrix of $\mathbf{M}$ corresponding to the dimensions of $\pmb{\mu}_y^c$ forms the dimension-reduced transformation.

\subsection{Normalization flow}

Normalization Flow (NF) is a deep generative model based on the principle of distribution transformation for continuous variables~\cite{rudin2006real}. 
If a latent variable $\mathbf{z}$ and an observation variable $\mathbf{x}$ are linked by an invertible 
transformation as $\mathbf{x} = f(\mathbf{z})$, their probability densities have the following relationship:

\begin{equation}
\label{eq:flow}
p(\mathbf{x}) = p(\mathbf{z})|\mathbf{J_{\mathbf{x}}}|,
\end{equation}

\noindent where the first term on the right hand side is the prior probability of the latent codes, which we set to be a standard Gaussian;
the second term $\mathbf{J_{\mathbf{x}}}=\det \frac{ \partial f^{-1}(\mathbf{x})}{\partial \mathbf{x}}$ is the determinant of the 
Jacobian matrix of $f^{-1}$, often known as the entropy term.
It has been shown that if $f$ is flexible enough, any complex distribution $p(\mathbf{x})$ can be transformed to 
a standard Gaussian~\cite{papamakarios2019normalizing}. Usually $f$ is implemented by a neural networks.

The NF model can be trained by maximum likehood with respect to the parameters of $f$. Given a set of training data $\{\mathbf{x}_i\}$,
the likelihood function can be written by:

\[
L(\bm{\theta}) = \sum_i \log p(\mathbf{x}_i) = \sum_i \big\{\log p(\mathbf{z}_i) + \log |\mathbf{J}_{\mathbf{x}_i}|\big\},
\]
\noindent where $\bm{\theta}$ denotes the parameters of the NF network. Since $p(\mathbf{z})$ is a Gaussian, 
the prior term in the objective can be easily computed. For the entropy term, however, a special structure design is required to ensure that the 
Jacobian matrix is tractable, e.g.,~\cite{dinh2016density,papamakarios2017masked}. The objective function can be optimized by any 
numerical optimizers, e.g., stochastic gradient descent (SGD).

\subsection{DNF: real deep LDA}

\begin{figure}[htb]
    \centering
    \includegraphics[width=0.7\linewidth]{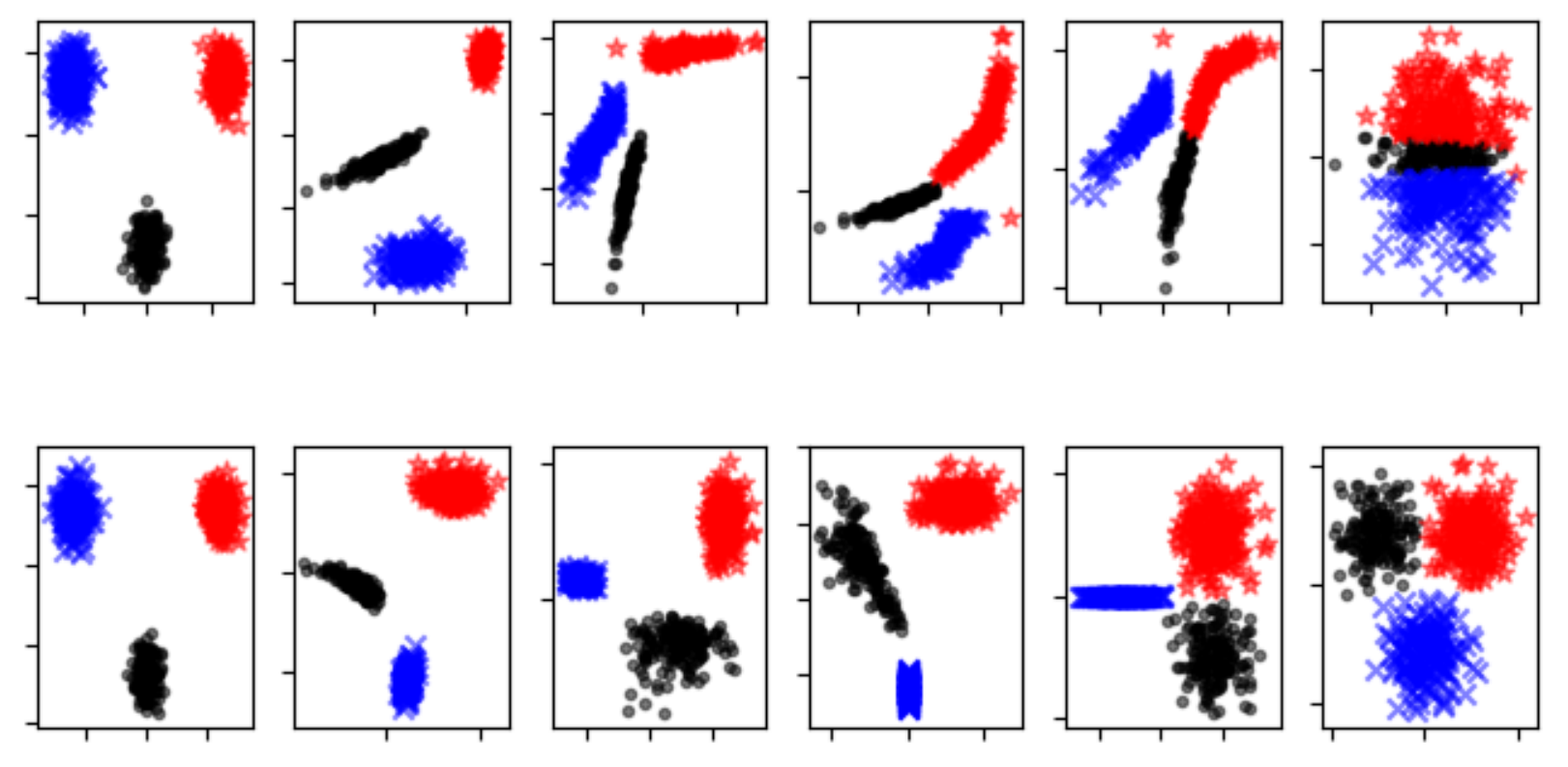}
    \vspace{-3mm}
    \caption{Normalization process of vanilla NF (top) and DNF (bottom) with  2-dimensional data sampled from 3 Gaussian components. Vanilla NF transforms 
    all the data to be a single Gaussian, while DNF transforms each class to an individual Gaussian.}
    \label{fig:flow-map-dnf.pdf}
\end{figure}

NF is a powerful tool for normalizing complex distributions, however it does not consider any class information and therefore may lose
the class structure during the projection, as shown in the top row of Fig.~\ref{fig:flow-map-dnf.pdf}.

To solve this problem, we extended NF to DNF, a discriminative normalization flow~\cite{cai2020deep}. It is simply a NF but each class 
owns its individual class mean:

\begin{equation}
\label{eq:dnf}
 p(\mathbf{x}; y(\mathbf{x})) = p(\mathbf{z}; y(\mathbf{x})) \Bigg|\det \frac{\partial f^{-1}(\mathbf{x})}{\partial \mathbf{x}}\Bigg|,
\end{equation}

\noindent where $p(\mathbf{z}; y(\mathbf{x}))$ is a Gaussian in the form of Eq. (\ref{eq:LDA-prior-z}). DNF can be trained in the same fashion as NF.
As shown in the bottom row of Fig.~\ref{fig:flow-map-dnf.pdf}, DNF projection preserves the class structure.

An interesting observation is that the likelihood function in LDA (Eq.(\ref{eq:ML-LDA})) and DNF (Eq.(\ref{eq:dnf})) are very similar.
The only difference is on the entropy term, which is based on a linear transformation $\mathbf{x} = \mathbf{M}^{-1}\mathbf{z}$ in LDA
and an invertible (may be nonlinear and complex) transformation $\mathbf{x}=f(\mathbf{z})$ in DNF. Besides the resemblance in the form of likelihood function, LDA and DNF
share the same assumption that the class distributions in the latent space are homogeneous Gaussians, and the training criterion are both based on
maximum likelihood. 
Essentially, the two models can be regarded fundamentally the same in the sense 
that both try to \textbf{\emph{explain data by standard Gaussians}}. Nevertheless, the nonlinear transformation of DNF substantially extend the range of
data that can be `explained', especially when the transformation is materialized by deep neural networks. For example, DNF is able to deal with
non-homogeneous and non-Gaussian class distributions.

\subsection{Subspace DNF}

Thanks to the resemblance between DNF and LDA, dimension reduction can be conducted following the same argument of Kumar~\cite{kumar1998heteroscedastic}. 
Specifically, we divide the latent space into a class-dependent part and a class-independent part, as does in the dimension-reduction LDA shown in Eq.(\ref{eq:split}),
and then train the DNF model as usual. By this setting, the class means are constrained in a subspace spanned by $\bm{\mu}^c_y$, so the model is
called \textbf{subspace DNF}. Compared to dimension-reduction LDA, subspace DNF performs a nonlinear dimension reduction.

\section{Related works}

Researchers have tried to combine deep neural networks and LDA, and call the resultant model `deep LDA'. For example, Stuhlsatz et al~\cite{stuhlsatz2012feature} pre-trained a deep neural networks
and then refined the net by Fisher discriminant criterion. The same idea motivated the work by Matthias et al.~\cite{dorfer2015deep}, where
the Fisher discriminant criterion was formulated as a function of the eigenvalues of $\bm{S}_W^{-1}\bm{S}_B$, and the model was trained in an end-to-end fashion. All of them focused on the Fisher
criterion superficially in order to mimic LDA, but that criterion is not necessarily better than cross entropy.


\section{Simulation experiment}

\begin{figure*}[htb!]
    \centering
    \includegraphics[width=1.0\linewidth]{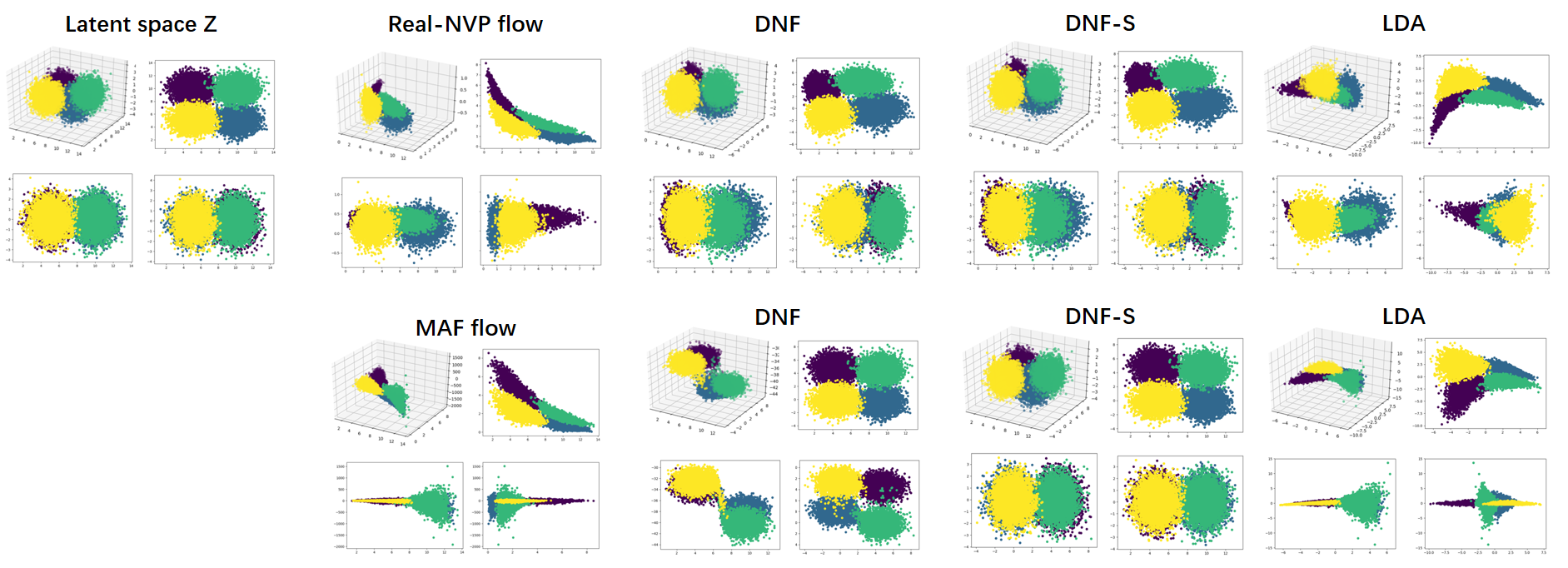}
    \vspace{-5mm}
    \caption{2D and 3D visualization for the simulation experiment. The first column shows the sample code $\bm{z}$; the second column shows the
    observation data $\mathbf{x}$ transformed from $\mathbf{z}$ by two NF models; the third to fifth columns show 
    the recovered latent code $\mathbf{z}$ by DNF, subspace DNF and LDA, respectively.}
    \label{fig:sim_data.png}
\end{figure*}

In this section, we compare LDA and DNF by a simulation experiment. 
We firstly sample 3-dimensional latent codes $\mathbf{z}$ of 4 classes, 
each class being an independent Gaussian. The class means span in two dimensions, and the third 
dimension is simply set to be zero for all the class means. We use 
two randomly initialized NF models to transformation $\mathbf{z}$ to  $\mathbf{x}$:
one is a Real NVP flow~\cite{dinh2016density} with 20 blocks and the other is a MAF flow~\cite{papamakarios2017masked} with 10 blocks. 
The latent codes and the observations 
generated by the two flows are shown in the first two columns in Fig.~\ref{fig:sim_data.png}. In each 
group of pictures, we show both the 3-D view and 2-D views from three angles. 
It can be seen that the distributions of the observation data generated by the two flows are rather complex.

Now we perform DNF, subspace DNF (DNF-S) and LDA to map the data $\mathbf{x}$ back to the latent codes $\mathbf{z}$,
as shown in column 3-5 in Fig.~\ref{fig:sim_data.png}. It is clearly to see that LDA 
fails to explain the data, as the latent codes do not distribute as the original 4 Gaussians. In 
contrast, DNF-S recovers the latent distribution almost perfectly. DNF can also find the clusters of the classes, 
however the class means may be scattered in the full 3-dimensional space. This is expected as there is not a 
subspace constraint in this model.

\section{Experiment on speaker recognition}

Modern speaker recognition is based on deep speaker embedding to extract speaker vectors and employs PLDA~\cite{Ioffe06} to score 
trials. In this experiment, we employ LDA and subspace DNF to perform dimension reduction for speaker vectors, in order to 
improve sample efficiency for the PLDA modeling. 

\subsection{Data and setting}

Three datasets were used in our experiments: VoxCeleb~\cite{nagrani2017voxceleb,chung2018voxceleb2}, SITW~\cite{mclaren2016speakers} and CNCeleb~\cite{fan2019cn}. 
VoxCeleb was used to train models (x-vector, PLDA, LDA, subspace DNF), while the other two were used for performance evaluation. We use x-vector model to produce speaker vectors. The model was created using the Kaldi toolkit~\cite{povey2011kaldi}, following the SITW recipe. 

We implemented the subspace DNF model with the MAF architecture~\cite{papamakarios2017masked}. It consists of 10 MAF blocks, 
and each block is an inverse autoregressive transformation. 
For more details of the MAF architecture, please refer to~\cite{papamakarios2017masked}, or the code we published online\footnote{https://github.com/Caiyq2019/Deep-generative-LDA}.
More details of the experimental system can be found in our previous work~\cite{cai2020deep}.

\subsection{Results}

\vspace{-2mm}
\begin{table}[htb!]
 \begin{center}
  \caption{EER(\%) results on SITW and CNCeleb.}
  \label{tab:eer}
  \scalebox{0.9}{
   \begin{tabular}{|l|l|c|c|}
   \hline
   \multicolumn{2}{|c|}{Model}             & \multicolumn{1}{c|}{SITW}   & \multicolumn{1}{c|} {CNCeleb}\\
   \hline
            Dimension Reduction           & System [Dim]        & PLDA    & PLDA \\
   \hline

            N/A                 & x-vector [512]     & 5.30    & 13.03\\
   \hline
            Linear              & LDA [512]          & 5.30    & 13.03 \\
                                & LDA [400]          & 4.65    & 12.28 \\
                                & LDA [200]          & 3.96    & 13.50 \\
                                & LDA [150]          & 4.07    & 14.37 \\
   \hline
           Nonlinear            & DNF-S [512]        & 3.66     &  11.82 \\
                                & DNF-S [400]        & \textbf{3.53} & \textbf{11.66} \\
                                & DNF-S [150]        & 4.62 & 12.83 \\
  \hline
\end{tabular}}
\end{center}
\end{table}
\vspace{-4mm}

Table~\ref{tab:eer} shows the results in terms of Equal Error Rate (EER). Firstly notice that on SITW, which 
is mostly an in-domain test, both LDA and subspace DNF achieve substantial performance improvement, compared to the original x-vector system. 
For LDA, the best performance is obtained when the dimension is reduced to 200. For subspace DNF, we can only reduce 
the dimension to 400, and most performance improvement is not due to dimension reduction, but the nonlinear normalization
provided by the full-dimension DNF~\cite{cai2020deep}. Nonetheless, the results show that a nonlinear dimension reduction 
can be achieved by subspace DNF, and it may lead to additional performance gain over nonlinear normalization.
On CNCeleb, which is an out-of-domain test, subspace DNF shows more advantage, demonstrating its better generalization capability. 

\begin{figure}[htb]
    \centering
    \includegraphics[width=1\linewidth]{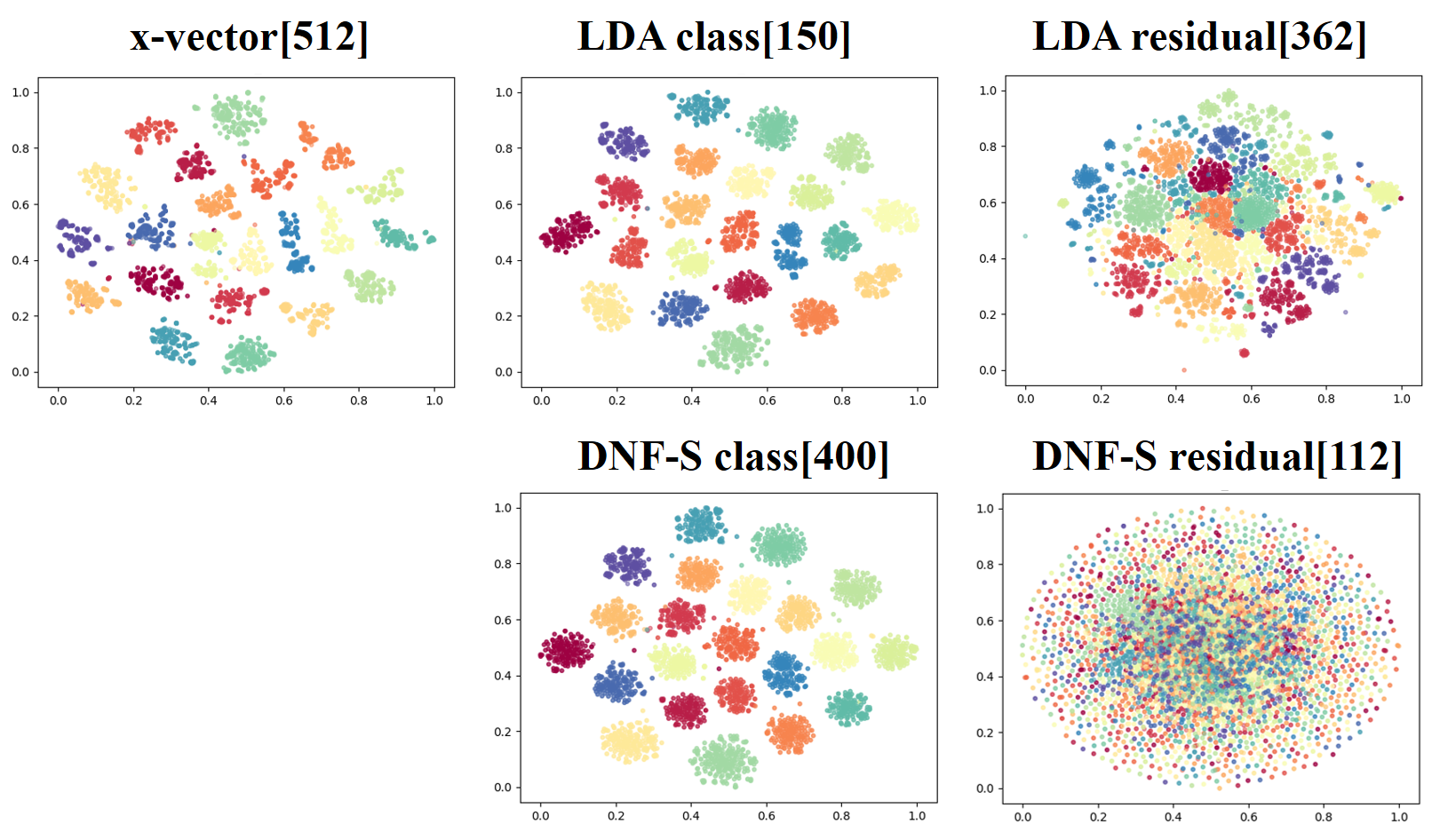}
    \vspace{-6mm}
    \caption{Distributions of the latent codes in the class space and the residual space.
    Each color represents a speaker, and each point represents a speaker vector. 
    The pictures are plotted by using t-SNE~\cite{maaten2008visualizing}.}
    \label{fig:dist.png}
\end{figure}

Fig.~\ref{fig:dist.png} shows the latent space corresponding to class-dependent dimensions (class space) and class-independent dimensions (residual space), 
derived by LDA and subspace DNF, respectively. We can see that the class information has been eliminated much better in the residual space 
derived by subspace DNF than by LDA, and the Gaussianity of the latent codes in the class space derived by subspace DNF is better than by LDA.

\vspace{-1mm}
\vspace{-2mm}

\section{Conclusions}

This paper interpret the discriminative normalization flow (DNF) as a deep generative LDA, and presents a subspace DNF that can perform nonlinear dimension reduction.
Our analysis shows that our model represents a natural nonlinear extension for LDA while keeping almost all its valuable features, e.g., data explanation and data generation.
Experiments on a simulation task and a practical speaker recognition task demonstrated that DNF and its subspace version are highly effective in modeling complex data and
retrieving low-dimensional representation for them.



\newpage
\setlength{\baselineskip}{11.5pt}
\bibliographystyle{IEEEbib}
\bibliography{refs}
\end{document}